\documentclass{article} %
\usepackage{iclr2023_conference,times}

\usepackage{amsmath,amsfonts,bm}

\def\eqref#1{equation~\ref{#1}}

\def\1{\bm{1}}

\DeclareMathAlphabet{\mathsfit}{\encodingdefault}{\sfdefault}{m}{sl}
\SetMathAlphabet{\mathsfit}{bold}{\encodingdefault}{\sfdefault}{bx}{n}

\usepackage{algorithm}
\usepackage{algpseudocode}

\usepackage{url}
\usepackage{booktabs}
\usepackage{graphicx}
\usepackage{fixltx2e}
\usepackage{multirow}
\usepackage{makecell}
\usepackage{color}
 \definecolor{mydarkblue}{rgb}{0,0.08,0.45}
\usepackage[colorlinks,citecolor=mydarkblue,urlcolor=mydarkblue,linkcolor=mydarkblue]{hyperref}
\usepackage{float}

\usepackage{amssymb}%
\usepackage{pifont}%
\newcommand{\cmark}{\ding{51}}%
\newcommand{\xmark}{\ding{55}}%

\definecolor{myRed}{rgb}{1,0.0,0}

\definecolor{myGreen}{rgb}{0.0,1.0,0}

\definecolor{myBlue}{rgb}{0,0.0,1.0}

\definecolor{myPink}{rgb}{1.0,0.0,1.0}

\definecolor{myYellow}{rgb}{0.5,0.5,0.0}

\title{Recitation-Augmented Language Models}

\author{
Zhiqing Sun$^{1,2*}$, Xuezhi Wang$^{1}$, Yi Tay$^{1}$, Yiming Yang$^{2}$, Denny Zhou$^{1}$\\
$^1$Google Research, Brain Team\\
$^2$Language Technologies Institute, Carnegie Mellon University
}

\newcommand{\palm}{PaLM}
\newcommand{\method}{RECITE}

\newcommand\blfootnote[1]{%
  \begingroup
    \renewcommand\thefootnote{}\footnote{#1}%
  \addtocounter{footnote}{-1}%
  \endgroup
}

\iclrfinalcopy %

\begin{document}

\maketitle

\blfootnote{* Work done during internship at Google.}

\begin{abstract}
We propose a new paradigm to help Large Language Models (LLMs) generate more accurate factual knowledge without retrieving from an external corpus, called RECITation-augmented gEneration (\method). Different from retrieval-augmented language models that retrieve relevant documents before generating the outputs, given an input, \method~first recites one or several relevant passages from LLMs' own memory via sampling, and then produces the final answers.
We show that \method~is a powerful paradigm for knowledge-intensive NLP tasks.
Specifically, we show that by utilizing recitation as the intermediate step, a recite-and-answer scheme can achieve new state-of-the-art performance in various closed-book question answering (CBQA) tasks.
In experiments, we verify the effectiveness of \method~on four pre-trained models (\palm, UL2, OPT, and Codex) and three CBQA tasks (Natural Questions, TriviaQA, and HotpotQA). Our code is available at \url{https://github.com/Edward-Sun/RECITE}.

\end{abstract}

\section{Introduction}

Large language models (LLMs) have achieved impressive in-context few-shot performance on knowledge-intensive NLP tasks \citep{brown2020language,rae2021scaling,hoffmann2022training,chowdhery2022palm}. For example, in open-domain question answering \citep{chen2017reading}, demonstrated by only a few examples of question-answer pairs, LLMs are able to answer arbitrary factoid questions \citep{joshi2017triviaqa,yang2018hotpotqa,kwiatkowski2019natural}.
Recent research \citep{guu2020retrieval,lewis2020retrieval,izacard2022few} shows that retrieval-augmentation can further improve LLMs' performance on knowledge-intensive tasks by conditioning the LLMs on retrieved relevant passages from an external corpus.

\begin{figure}[H]
    \centering
    \includegraphics[width=0.9\linewidth]{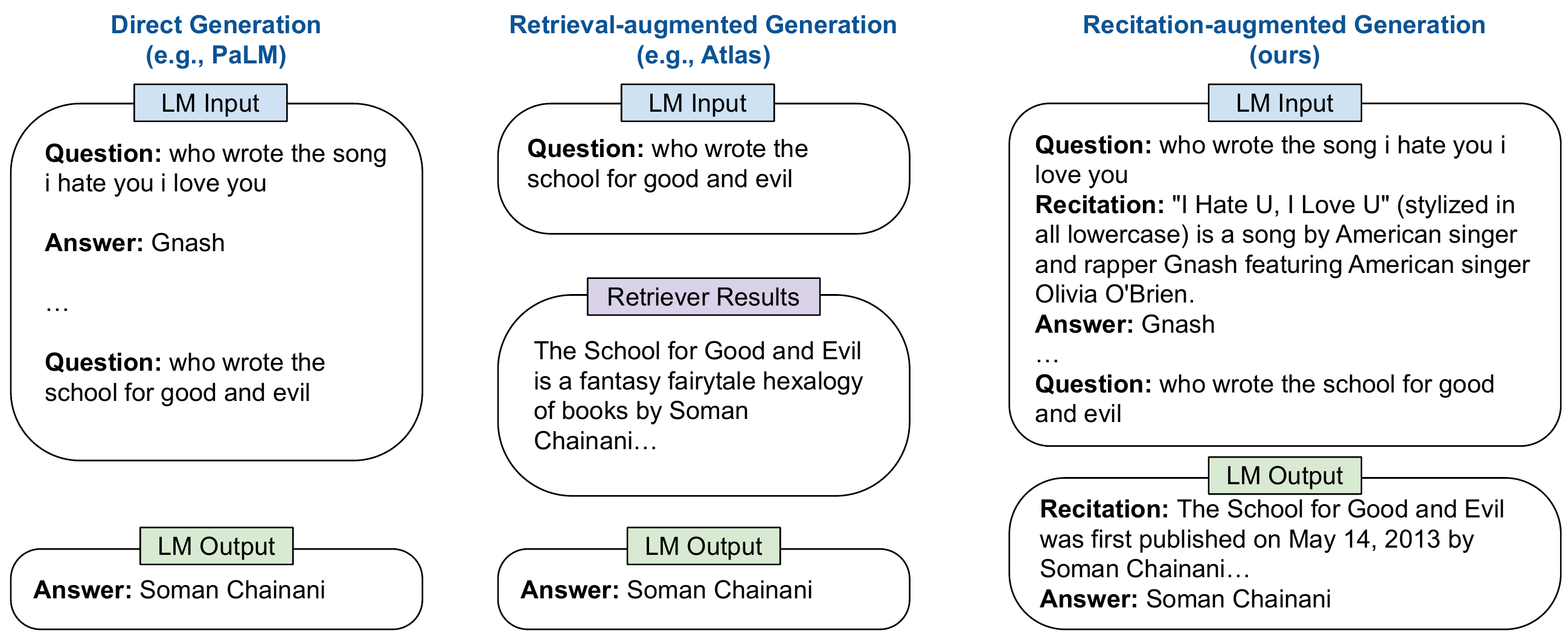}
    \caption{Illustration of evaluating (few-shot) open-domain question answering with (closed-book) direct generation \citep{chowdhery2022palm}, (open-book) retrieval-augmented generation \citep{izacard2022few}, and (closed-book) recitation-augmented generation (ours).}
    \label{fig:comparison}
\end{figure}

This paper proposes a new paradigm to help LLMs generate more accurate factual knowledge without retrieving from an external corpus, called RECITation-augmented gEneration (\method), wherein
we tackle knowledge-intensive NLP tasks by first reciting relevant information and then generating the outputs.
Such a two-step paradigm decomposes the original knowledge-intensive task into two sub-tasks: knowledge-recitation and task-execution, where the former can be regarded as a form of intermediate knowledge retrieval step (from the model weights), while the latter is the execution step that produces the final outputs.

The motivation of introducing an additional knowledge-recitation step comes from our observation that while few-shot prompting can help LLMs execute specific NLP tasks, these tasks are usually not in a similar form as the original causal language modeling pre-training objective. This hinders LLMs from effectively reciting knowledge from their memory \citep{carlini2021extracting}.
Consider a student taking a closed-book exam that contains knowledge-intensive questions, for example, \textbf{``what is the tenth decimal of $\boldsymbol{\pi}$?''}. They typically cannot directly answer this question because in studying stage (in analogy to the language modeling pre-training stage for LLMs), it is highly unlikely that they would read ``the tenth decimal of $\pi$ is 5''. However, there can be some sentences like ``the first $N$ digits of $\pi$ are 3.14159 26535...'' existing in the textbook that can be recited by the student. Therefore, a student can possibly answer this question in a recite-and-answer scheme: \textbf{``The first 10 digits of $\boldsymbol{\pi}$ are 3.14159 26535. So the answer is 5''}.
Here, the knowledge-recitation step can serve as an intermediate step that mimics the language modeling pre-training task, and thus better helps the LLM to generate factual knowledge.

We verify the effectiveness of our recitation-augmented generation on few-shot Closed-Book Question Answering (CBQA) tasks (referred as \textbf{recite-and-answer} in the CBQA context), as illustrated in Figure~\ref{fig:comparison}.
CBQA is an attractive open-domain QA task in that a fully parameterized LM can generate answers directly without an external corpus or separate retrieval models \citep{roberts2020much}.
We show that the proposed recite-and-answer scheme is an effective method for CBQA and compatible with other techniques for boosting few-shot performance of LLMs. We also show that, in addition to improving the few-shot in-context learning performance of \method-enhanced LLM, fine-tuning the pre-trained LLMs on synthetic generated question-passage pairs can further improve the recitation performance and lead to a better downstream QA accuracy. 

Experiments on four large language models (\palm~\citep{chowdhery2022palm}, UL2 \citep{tay2022unifying}, OPT \citep{zhang2022opt}), and Codex \citep{chen2021evaluating} show that a recite-and-answer scheme can improve performance on various types of CBQA tasks, including Wikipedia-based single-hop QA (Natural Questions, \citealt{kwiatkowski2019natural}), trivia questions (TriviaQA, \citealt{joshi2017triviaqa}), and Wikipedia-based multi-hop QA (HotpotQA, \citealt{yang2018hotpotqa}).

\section{Related Work}

\subsection{Open-domain question answering}

Open-domain question answering \citep{prager2007open} refers to the task of generating answers for arbitrary context-free questions. In the open-book setting, it is typically assumed that the QA model can find the answer in an external corpus, e.g., Wikipedia  \citep{chen2017reading,izacard2021leveraging} or web pages \citep{lazaridou2022internet}. This is in analogy as taking an open-book exam where students can search over an external knowledge corpus. The standard pipeline \citep{chen2017reading,izacard2021leveraging,izacard2020distilling} usually consists of a learnable or non-learnable document retriever module and a learnable neural network-based reader module.

In the closed-book setting, the QA model is not allowed to access any external knowledge, and needs to store all the knowledge in its parameters.
It has been recently observed that large-scale pre-trained language models \citep{devlin-etal-2019-bert,radfordimproving,yang2019xlnet} can internalize a sort of implicit ``knowledge base'' after pre-training \citep{petroni2019language,jiang2020can,talmor2020olmpics}. \citet{roberts2020much} show that after fine-tuning on open-book question-answer pairs, T5 \citep{raffel2020exploring} can answer a large portion of knowledge-intensive questions. This is similar as taking a closed-book exam.
However, \citet{lewis2021question} found that the high performance is mainly due to training set question memorization.
\citet{wang2021can} also found that it is still challenging for relatively small-scale pre-trained language models like RoBERTa \citep{liu2019roberta} or GPT-2 \citep{radfordlanguage} to answer closed-book questions.

In this work, we focus on evaluating the CBQA performance of large language models (LLMs) in the few-shot setting, which ideally minimizes the bias of train-test overlapping \citep{liu2021challenges}.
We propose a recite-and-answer scheme, which is similar to a student first recite the factoid knowledge about the question, and then answer the question.

\subsection{In-context few-shot learning}

Large language models (LLMs) such as GPT-3 \citep{brown2020language} have the surprising ability to do in-context learning, where the model learns to do new tasks simply by being prompted a few exemplars. The LLMs learn from these exemplars without being explicitly pre-trained for in-context learning and without any gradient updates or fine-tuning. Recent study showed that such ability improves with the scaling of both model size \citep{brown2020language,rae2021scaling,chowdhery2022palm} and number of tokens for training \citep{hoffmann2022training}. When evaluated on knowledge-intensive question answering tasks, these models are usually evaluated in the closed-book setting, where the factoid knowledge are completely stored in the model parameters of dense LLMs.

Recently, Atlas \citep{izacard2022few} shows that for knowledge-intensive NLP tasks, a relatively lite model with retrieval augmentations can achieve similar or even better performance through few-shot fine-tuning, which proves that memorization can be decoupled from generalization in LLMs. In contrast, we show that still a large underestimated amount of knowledge can be retrieved from LLMs' model weights through better-designed prompting. 

\subsection{Rationale-augmented reasoning}

\cite{ling2017program} pioneer the work of solving math word problems by generating step-by-step human-readable  solutions described by natural language and math equations before the final answer. That is fundamentally different from other works which directly generate the final answers or use  formal languages. e.g. equations only,  to illustrate the intermediate solving steps \citep{roy2016equation, amini2019mathqa, chen2019neural}. \cite{cobbe2021training} extend \citep{ling2017program} by constructing a much larger dataset to finetune a pre-trained large language model to solve math word problems and a parameterized ranker is trained to rank candidate solutions to improve the solving rate. \cite{wei2022chain} propose chain-of-thought prompting which combines the idea of  natural language rationales \citep{ling2017program, cobbe2021training} with few-shot prompting \citep{brown2020language}. 

In this work, instead of generating a chain of thought for multi-step reasoning questions, we decompose the process of answering a knowledge-intensive question into two steps: recite the relevant knowledge stored in the model parameters, and then answer the question.

\subsection{Memorization in large language models}

Recent study shows that large language models can memorize its training data, and generate texts from training data given certain prompts \citep{carlini2021extracting,carlini2022quantifying,zhang2021counterfactual,kharitonov2021bpe,thakkar2020understanding,carlini2019secret,tirumala2022memorization}. Most related to our work, \citet{carlini2022quantifying} found that the memorization ability of LLMs significantly grows as the model capacity increases, the number of times an example has been duplicated, and the number of tokens of context used to prompt the model.
While these works mainly analyze the fundamental properties of memorization in the exact setting, where exactly $N$ tokens are used as the prompt to reproduce the suffix of the prompt, our work relies on ``fuzzy memorizaiton'', where the prompts tend to not be exactly the same as the training data, but still improve the memorization accuracy.

The proposed recitation-augmented generation idea is also related to the line of work on utilizing Transformer memory as an information retrieval model \citep{tay2022transformer} and self-talk models for commonsense reasoning \citep{shwartz2020unsupervised,liu2022generated}. \citet{zhuang2022bridging,wang2022neural,zhou2022ultron} proposed to augment documents at indexing time with a number of generated queries. \citet{bevilacqua2022autoregressive} proposed to directly generate n-grams grounded in one or multiple documents with constrained decoding.

\section{Learning to Recite for Closed-book Question Answering}

\begin{figure}[t]
\centering
\includegraphics[width=0.8\linewidth]{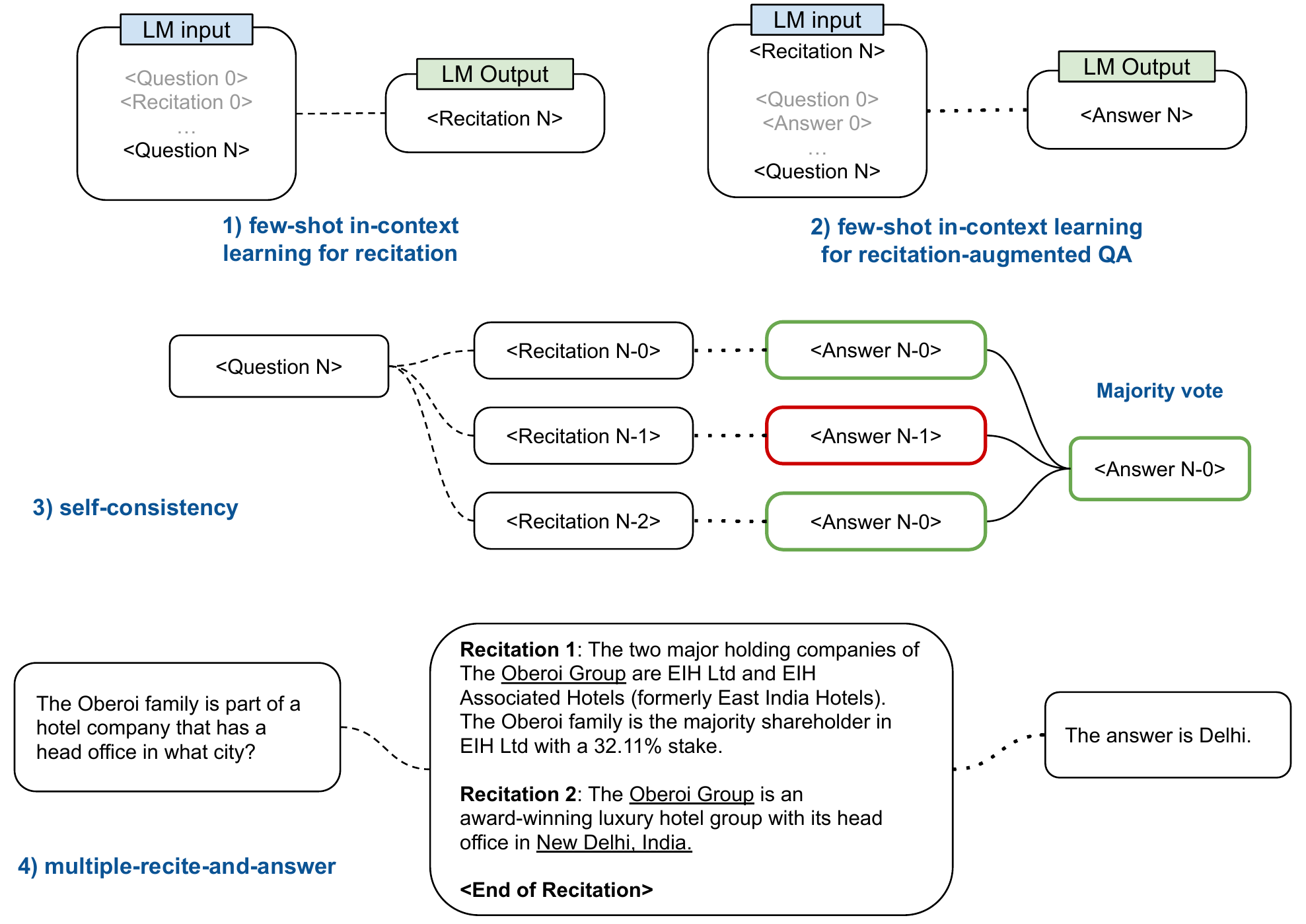}
  \caption{Illustration of prompt-based in-context learning for recitation generation, recitation-augmented question answering, self-consistency ensembling, and multiple-recite-and-answer for multi-hop questions (Sec.~\ref{sec:prompting}).
  In multiple-recite-and-answer scheme, the latter recitaiton can utilize the information from the previous ones, such as ``Oberoi Group'' in this case.
  The prompts for self-consistency and multi-hop recite-and-answer are dropped for brevity.}
 \label{fig:multi_hop}
\end{figure}

The goal of this paper is to mimic a human's ability to recite relevant factoid knowledge \citep{mcdaniel2009read} before answering knowledge-intensive questions, such that these questions can be answered more accurately.
In the following we describe our recite-and-answer scheme for few-shot closed-book question answering (CBQA), which consists of two components: (1) a evidence-recitation module for reciting relevant passages, and (2) a question-answering module for generating answers given the recited evidence. Notice that in this paper, we focus on few-shot setting, in which we assume only a few question-answer demonstrations are provided.
In Natural Questions \citep{kwiatkowski2019natural} benchmark, since the questions are from queries issued to the Google search engine by multiple users, and thus can be regarded as unannotated data, we further assume that we have top-retrieved Wikipedia pages for these questions. The paragraphs in these top-retrieved Wikipedia pages will be used to generate synthetic paired question-recitation data for fine-tuning the LM (described in Section~\ref{sec:fine-tuning}).

\subsection{Prompt-based recite-and-answer for question-answering}\label{sec:prompting}

\paragraph{Recitation-augmented question answering}
We start with single-hop question answering \citep{kwiatkowski2019natural,joshi2017triviaqa}, where the answers are usually supported by a specific document in the corpus, which is sometimes referred as evidence \citep{joshi2017triviaqa}. Different from chain-of-thought prompting \citep{wei2022chain} where a rationale is directly generated to explain the generated answer \citep{joshi2017triviaqa,narang2020wt5,lampinen2022can}, we propose to first recite a passage about the question, and then answer the question based on the recitation.

We propose a prompt-based learning-to-recite scheme by leveraging the LLM's in-context learning ability \citep{brown2020language}. We prompt the LLM with paired exemplars of questions and recited evidences, and the LLM can learn in an in-context manner to generate a recitation for an arbitrary question.
To perform recitation-conditioned few-shot question answering, we append the recited passages at the beginning of the original question-answer exemplars as a single prompt, and then generate the final answer (Step 1 \& 2 in Figure~\ref{fig:multi_hop}).

\paragraph{Self-consistency ensemble}

The factual knowledge about a question can appear in several places in the language model's training corpora. For example, the fact of ``Queen Elizabeth II opened the London Bridge on 17 March 1973'' can appear in both Wikipedia page ``London Bridge'' and page ``March 1973'', so it is highly likely that there exists knowledge from different articles that could lead to the same, correct answer. With this motivation, we argue that similar to multi-step reasoning in chain-of-thought, recitation-augmented question answering can also benefit from the self-consistency technique with multiple-path decoding \citep{wang2022self}.
Specifically, given an arbitrary question, we first use top-$k$ sampling to independently generate a few recitations, and then greedy decode the answer of the question based on the sampled recitations. Finally, we determine the optimal answer by taking a plurality/majority vote (Step 3 in Figure~\ref{fig:multi_hop}).

\paragraph{Multiple-recite-and-answer for multi-hop question-answering}

Multi-hop question answering requires the QA system to find and reason over multiple supporting documents. However, the nature of recitation restricts us to recite passages from one article at a time.
In order to apply the recite-and-answer scheme to solve multi-hop questions, we introduce multiple-recite-and-answer scheme (Step 4 in Figure~\ref{fig:multi_hop}), that is, given the multiple-hop question, we use the prompt words such as ``Recitation 1'' and ``Recitation 2'' to elicit the LLM to generate recitation passages on different topics. Since the multiple recited passages are generated in one-pass from the LLM decoding sequentially, the generation of later passages can effectively utilize the information both in the original question and the previous recited ones.
Our multiple-recite-and-answer scheme for multi-hop question-answering is also compatible with the self-consistency technique, by applying top-$k$ sampling when generating multiple recitations and performing majority voting for the final answers.

\begin{figure}[t!]
    \centering
    \includegraphics[width=0.8\linewidth]{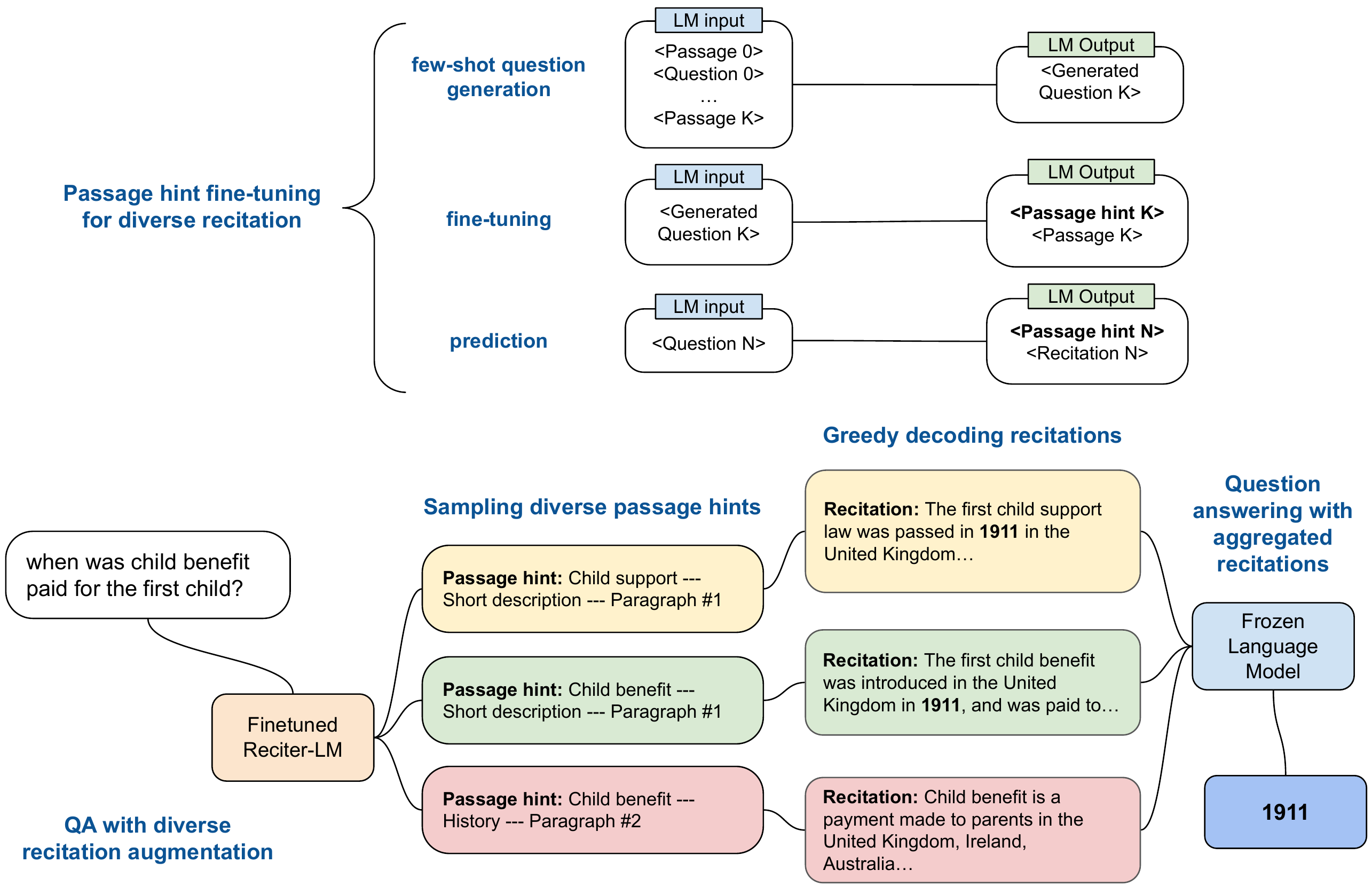}
    \caption{Illustration of question answering with diverse recitation and the corresponding few-shot question generation and fine-tuning processes.}
    \vspace{-0.1in}
    \label{fig:self-consistency}
\end{figure}

\subsection{Passage hint-based diversified recitation with fine-tuning} \label{sec:fine-tuning}

\paragraph{Passage hint-based diversified recitation}

While the sampling-based recitation and self-consistency improves the robustness of recite-and-answer method, one argument for its inefficiency is that if the evidence-recitation module samples the wrong facts about the question, the question-answering module will not be able to figure it out and tend to generate the wrong answer. Therefore, on the one hand, we need to use a low sampling temperature to avoid generating recitations with wrong facts, on the other hand, we want to make sure the sampled recitations have enough diversity.

To tackle such a dilemma, we propose \textit{passage hint-based diversified recitation}. We observe that in well-formed text knowledge bases, such as Wikipedia, we can usually find a unique passage hint for each passage, by concatenating the section titles and the in-section order of each passage. For example, the passage hint of the second passage in Section 5.2 ``Enforcement'' of Wikipedia page ``Child support'' would be ``Child support --- Compliance and enforcement issues --- Enforcement --- Paragraph \#2''. In passage hint-based diversified recitation, we first use \textit{sampling} to generate a diverse set of passage hints, and then use \textit{greedy decoding} to ensure the factual accuracy of the contents in each passage.

Since each passage hint corresponds to a unique passage, we can first de-duplicate the passage hints and then generate the full passages to get more diverse recitation passages. Furthermore, as the recited passages are less likely to be similar due to unique passage hints, inspired by recent progress on question-answering with multiple retrieved passages \citep{izacard2021leveraging}, we use aggregated diverse recitations as a single context, and generate the answer with a few more question-answer pair demonstrations. 
Figure~\ref{fig:self-consistency} (lower) illustrates the recite-and-answer scheme with passage hint-based diversified recitation.

\paragraph{Fine-tuning on few-shot generated questions}

We found that although the training data of many existing LLMs \citep{devlin-etal-2019-bert,chowdhery2022palm} contains the Wikipedia corpora, which are usually regarded as the factoid documents for knowledge-intensive question answering tasks \citep{joshi2017triviaqa,kwiatkowski2019natural}, the section titles are usually not explicitly included in training. This makes the pre-trained LLM hard to discover the mapping from the question to the passage hint, and to the full passage merely by few-shot prompting.

To address this issue, we propose an additional fine-tuning stage to adapt LLMs to learn such mappings. Assuming we have access to not only a few question-answer pairs, but also the top-retrieved Wikipedia pages for queries issued to the Google search engine by multiple users \citep{kwiatkowski2019natural}, we can use few-shot prompting to generated synthetic question-hint-passage pairs and then finetune the LLMs on the generated data.

Specifically, we use the ground-truth evidence and question pairs as the prompt, and generate new questions with in-context learning for randomly sampled passages from Wikipedia pages. Next, based on the few-shot generated questions, we train the LLM to predict the original passage hint, as well as the passage content.
Figure~\ref{fig:self-consistency} (upper) illustrates the whole process of passage hint fine-tuning.

\section{Experiments}

In this section, we report empirical evaluations of our proposed \method~with recite-and-answer schemes on a diverse set of few-shot closed-book question answering tasks and different language models with varying scales.

\subsection{Experimental setup}

\subsubsection{Evaluation Datasets}

\paragraph{Natural Questions}

Natural Questions \citep{kwiatkowski2019natural} consists of questions aggregated from the Google search engine and the answers from the Wikipedia page in the top $5$ search
results. We treat it as a single-hop question answering task.
Since Natural Questions contains the so-called ``long answer'' annotations, which is a whole HTML bounding box containing enough information to infer the answer, we directly utilize the ``long answer'' as the ground-truth recitation exemplars in our prompt (Sec.~\ref{sec:prompting}).
In order to make a direct comparison with recent LLMs \citep{chowdhery2022palm,izacard2022few}, we evaluate our methods in 5-shot and 64-shot settings.

\paragraph{TriviaQA}

TriviaQA dataset \citep{joshi2017triviaqa} is constructed by collecting Trivia enthusiast authored question-answer pairs and their retrospectively collected evidence. Since there is no obvious way to collect a ``long answer'' in the retrospective evidence documents (the exact appearances of the answer may contain enough information to infer the answer), we evaluate TriviaQA in the single-hop 5-shot setting, and manually compose the recitation passage from Wikipedia for 5 randomly sampled training questions. The concrete prompt can be found in the appendix.

\paragraph{HotpotQA}

HotpotQA \citep{yang2018hotpotqa} is designed to explicitly test QA systems' ability to perform multi-hop reasoning. It is collected by explicitly composing questions requiring reasoning about multiple supporting context documents. Following \citet{wang2022rationale}, we evaluate HotpotQA as a multi-hop question answering task in the 4-shot setting. But instead of chain-of-thought prompting as in \citep{wang2022rationale}, we use multiple-recite-and-answer (Sec.~\ref{sec:prompting}) to achieve multi-step reasoning. We also provide the concrete prompt in the appendix.

\paragraph{Metrics}

We calculate the Exact Matching (EM) and F1 scores for the normalized answers, while the specific text normalization applied on each dataset can be slightly different.

\subsubsection{Pre-trained language models}

We evaluate the effectiveness of RECITE on four langauge models: \palm, UL2 \citep{tay2022unifying}, OPT \citep{zhang2022opt}, and Codex \citep{brown2020language,ouyang2022training,chen2021evaluating}. Due to the space limit, the detailed descriptions of them are provided in Appendix \ref{sec:lms}.

\subsection{Experiments}

We use the test split for all tasks if the test split is available and has labels for evaluation, otherwise we use the dev split. In addition, TriviaQA and HotpotQA are too large to run large language models on, so we used the first 1,024 data points for evaluation.

\begin{table}[t]
\caption{Performance comparison on Natural Questions (NQ), TriviaQA, and HotpotQA. The number of shots for each task are mentioned in parenthesis.}
\small
\setlength\tabcolsep{2pt}
    \centering
    \begin{tabular}{l | l c c c c}
    \toprule
    & & \palm-62B & UL2-20B & OPT-30B & Codex-002\\
    & & EM / F1 & EM / F1 & EM / F1 & EM / F1\\
    \midrule
    \multirow{4}{*}{NQ}
    &\multirow{2}{*}{Standard-prompting (direct)}& 25.76 / 36.47\textsubscript{(5)} & 10.16 / 20.17\textsubscript{(5)}& \multirow{2}{*}{14.97 / 22.93\textsubscript{(5)}} & \multirow{2}{*}{31.45 / 44.75\textsubscript{(5)}}\\
    & & \ 28.98 / 40.13\textsubscript{(64)} & \ 12.70 / 21.97\textsubscript{(16)}\\
    & \multirow{2}{*}{Recite-and-answer (20-path)} & \textbf{28.70} / \textbf{39.76}\textsubscript{(5)} & \textbf{14.16} / \textbf{23.13}\textsubscript{(5)}  & \multirow{2}{*}{\textbf{17.84} / \textbf{26.74} \textsubscript{(5)}} & \multirow{2}{*}{\textbf{35.84} / \textbf{49.12}\textsubscript{(5)}}\\
    & & \ \textbf{31.34} / \textbf{42.48}\textsubscript{(64)} & \ \textbf{14.94} / \textbf{24.29}\textsubscript{(16)}  \\

    \midrule
    \multirow{2}{*}{TriviaQA} &Standard-prompting (direct)& 65.38 / 71.85\textsubscript{(5)} & 48.73 / 54.32\textsubscript{(5)}& 45.90 / 50.68\textsubscript{(5)} & 81.84 / 86.09\textsubscript{(5)}\\
    &Recite-and-answer (20-path) & \textbf{65.84} / \textbf{72.10}\textsubscript{(5)} & \textbf{53.42} / \textbf{58.69}\textsubscript{(5)}  & \textbf{49.02} / \textbf{54.22}\textsubscript{(5)} & \textbf{83.50} / \textbf{88.03}\textsubscript{(5)}\\
    
    \midrule
    \multirow{3}{*}{HotpotQA} & Standard-prompting (direct)& 20.51 / 28.90\textsubscript{(4)} & 16.99 / 24.99\textsubscript{(4)}& 16.70 / 25.21\textsubscript{(4)} & 28.32 / 39.03\textsubscript{(4)}\\
    & Chain-of-thought (20-path)& 23.73 / 32.80\textsubscript{(4)} & 17.68 / 24.87\textsubscript{(4)}& 16.89 / 24.03\textsubscript{(4)} & 34.38 / 45.50\textsubscript{(4)}\\
    & Recite-and-answer (20-path) & \textbf{26.46 / 35.67}\textsubscript{(4)} & \textbf{19.04} / \textbf{27.32}\textsubscript{(4)}  & \textbf{17.77} / \textbf{26.58}\textsubscript{(4)} & \textbf{37.11} / \textbf{48.37}\textsubscript{(4)}\\
    \bottomrule
    \end{tabular}
    \vspace{-0.1in}
    \label{tab:performance_nq}
\end{table}

\subsubsection{Prompt-based results}

We report the single-hop closed-book question answering (CBQA) evaluation results on Natural Questions (NQ) and TriviaQA and the multi-hop CBQA evaluation results on HotpotQA. In Tab.~\ref{tab:performance_nq}, we report the results with prompt-based in-context learning and self-consistency.

From the tables, we can see that the proposed recite-and-answer scheme can significantly improve the CBQA performance on both datasets with various pre-trained language models. While the performance improvements on NQ is more consistent across different language models, we find that the improvements from recite-and-answer is more significant on smaller language models on TriviaQA. Our hypothesis is that the Trivia-style question usually contains more contextual information in the question, thus weakened the effectiveness of recitation for strong LLMs like \palm.

Besides, we can see that the recite-and-answer scheme can outperform the chain-of-thought prompting performance on the multi-hop reasoning task. Interestingly, we also find that for LLMs that have large gains from chain-of-thought (i.e., \palm), they also have large improvements from recite-and-answer.

\subsubsection{Results of passage hint-based diversified recitation}

For Natural Questions dataset, since it has the collection of top-retrieved Wikipeida pages corresponding to the unannotated queries issued to the Google search engine, we additionally report the diversified recitation results of fine-tuned \palm~model in Tab.~\ref{tab:performance_nq_finetune}.
From the table, we find that diversified recitation can further improve the performance of \palm~on the NQ dataset.
\begin{table}[t]
\caption{Performance comparison of \palm-62B on Natural Questions (NQ) dataset with standard-prompting, recite-and-answer with self-consistency sampling, and recite-and-answer with diversified recitation. The number of shots for each task are mentioned in parenthesis.}
\small
    \centering
    \begin{tabular}{l c c}
    \toprule
    & EM / F1\textsubscript{(5)} & EM / F1\textsubscript{(64)}\\
    \midrule
    Standard-prompting (direct)& 25.76 / 36.47
    & \ 28.98 / 40.13\\
    Recite-and-answer (20-path) & 28.70 / 39.76
    & \ 31.34 / 42.48\\
    Recite-and-answer w/ diversified recitation (20-path) & \textbf{32.20} / \textbf{44.02}
    & \ \textbf{33.23} / \textbf{45.29}\\
    \bottomrule
    \end{tabular}
    \vspace{-0.1in}
    \label{tab:performance_nq_finetune}
\end{table}

\subsection{Analysis}

\begin{figure}[H]
\centering
\begin{minipage}{.5\textwidth}
  \centering
  \includegraphics[width=.7\linewidth]{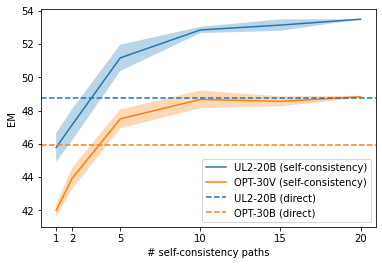}
\end{minipage}%
\hfill
\begin{minipage}{.5\textwidth}
  \centering
  \includegraphics[width=.7\linewidth]{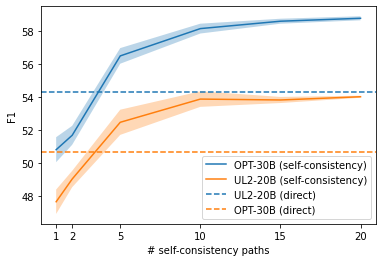}
\end{minipage}
\vspace{-0.1in}
\caption{TriviaQA EM/F1 on OPT-30B and UL2-20B with different \# of self-consistency paths.}
\label{tab:analysis_trivia}
\end{figure}

\subsubsection{On the number of self-consistency paths}

We analyze how the number of passages recited would affect the performance of recite-and-answer under the self-consistency setting.
Due to the costly inference of LLMs, we first sample up to $k=20$ recitation passages, and then apply self-consistency to a randomly selected subset of recitations to simulate less paths.
For each number of self-consistency paths, we evaluate the randomly selected subsets five times and report the mean and standard deviation.
We conduct an analysis on OPT-30B and UL2-20B on the TriviaQA dataset and report the results in Fig.~\ref{tab:analysis_trivia}. We can see that sampling more recitation passages tends to improve the recite-and-answer performance, while less randomness is observed with more self-consistency paths.

\subsubsection{On the robustness of few-shot exemplars}

A well-known problem of in-context few-shot learning is its instability to the choices of exemplars and their orders \citep{zhao2021calibrate}. We evaluate the robustness of standard prompting and our recite-and-answer method with 5 random seeds and report the mean and standard deviation of UL2 model running on the TriviaQA dataset in Tab.~\ref{tab:robustness_evaluation}. The 5-shot exemplars are randomly sampled and shuffled for each seed. From the table, we can see that with recitation sampling, recite-and-answer exhibits similar robustness (in terms of small performance deviation) as standard prompting under different random seeds and numbers of self-consistency paths. The overall gains by recite-and-answer are significant compared to standard prompting regardless of the choice of few-shot exemplars.

\begin{table}[t]
\begin{center}
    \begin{minipage}{0.8\textwidth}
\caption{Natural Questions (NQ) results with different context passages.}
    \centering
    \begin{tabular}{l c c}
    \toprule
    & UL2-20B\textsubscript{(5)} & Codex-002\textsubscript{(5)} \\
    & EM / F1 & EM / F1\\
    \midrule
    No passage& 10.16 / 20.17 & 31.45 / 44.75 \\
    \midrule
    Ground-truth passage & 41.02 / 55.73 & 49.32 / 64.32 \\
    BM25-Retrieval (Top-$1$) & 16.31 / 27.66 & 33.20 / 47.45 \\
    LM-Recitation\textsubscript{(5)} (20-path) & 14.16 / 23.13 & 35.84 / 49.12 \\
    \bottomrule
    \end{tabular}
    \vspace{-0.05in}
    \label{tab:context_ablation}  
\end{minipage}
\setlength\tabcolsep{3.0pt}
\end{center}
\setlength\tabcolsep{3.0pt}
\begin{minipage}{0.5\textwidth}
\caption{Per-question error analysis on TriviaQA.}
    \centering
    \begin{tabular}{l c c}
    \toprule
    & UL2-20B\textsubscript{(5)} & OPT-30B\textsubscript{(5)}\\
    \midrule
    Hits@Majority & 53.42\% & 49.02\% \\
    \midrule
    Not Recit. & 21.09\% & 22.27\% \\
    Hits@20-Recit. & 5.66\% & 8.01\% \\
    Hits@20-Path & 19.82\% & 20.07\% \\
    \bottomrule
    \end{tabular}
    \vspace{-0.05in}
    \label{tab:error_analysis}  
\end{minipage}
\begin{minipage}{0.5\textwidth}
\caption{Per-path error analysis on TriviaQA.}
    \centering
    \begin{tabular}{c c c c}
    \toprule
    Recit. & Ans. & UL2-20B\textsubscript{(5)} & OPT-30B\textsubscript{(5)}\\
    \midrule
    \cmark & \cmark & 33.60\% & 30.06\% \\
    \midrule
    \cmark & \xmark & 7.87\% & 9.79\% \\
    \xmark & \cmark & 12.10\% & 12.57\% \\
    \xmark & \xmark & 46.44\% & 47.58\% \\
    \bottomrule
    \end{tabular}
    \vspace{-0.05in}
    \label{tab:per_path_error_analysis}  
\end{minipage}
\vspace{-0.1in}
\end{table}

\subsubsection{Recitation v.s. Retrieval v.s. Ground-truth}

One may ask without the external corpus, whether the quality of recited passages with LLMs is better than simple retrieval models, e.g., BM25 \citep{robertson2009probabilistic}\footnote{We use the pre-indexed ``enwiki-paragraphs'' corpus in the pyserini package (\url{https://github.com/castorini/pyserini}), which is originally designed for BERTserini \citep{yang2019end}.}. To answer this question, we evaluate the few-shot question-answering performance of UL2 and Codex on three kinds of context passages: retrieval, recitation, and ground-truth. We report the results on first 1024 validation examples in Natural Questions (NQ) dataset, since it is the only dataset that contains the ``long answer'' annotation that can be regarded as ground-truth context passage. From Tab.~\ref{tab:context_ablation}, we can see that the classic retrieval model, i.e., BM25, is still a very strong baseline for collecting information from the corpus. Nonetheless, compared to BM25, our recite-and-answer still achieves a quite competitive  performance via generation only and without  using any external corpus.
Besides, we find that stronger models (i.e., Codex) tend to benefit more from the the model's own recitation than BM25 retrieved context.

\subsubsection{Error analysis}

We perform an error analysis on the 1024 evaluation examples in the TriviaQA dataset. We classify the errors into three categories: 1) Not Recit., i.e., the correct answer is not recited in any of the 20 recited passages in self-consistency. 2) Hits@20-Recit., i.e., the correct answer can be found in one of the recited passage, but does not appear in the QA module's outputs. 3) Hits@20-Path, i.e., the correct answer is one of the final outputs of the 20 self-consistency paths, but it does not have the majority votes. The correct final answer is marked as Hits@Majority (i.e., Exact Matching). An algorithmic description is given in Algo.~\ref{alg:error}.
We report the results of UL2-20B and OPT-30B in Tab.~\ref{tab:error_analysis}. We can see that ``No Recit'' and ``Hits@20-Path'' account for the majority of the errors, meaning that the QA module performs quite well (if the correct answer appears in one of the recitation passages, it will be extracted by the QA module in most of the cases), and the main bottleneck still lies in the recitation quality and answer aggregation strategies.

We also perform a per-path error analysis, i.e., how many questions can be answered correctly (or not) when the recitation exactly contains (or not) the answer tokens. The results are shown in Tab.~\ref{tab:per_path_error_analysis}. We can see that around $7\% \sim 10\%$ questions have the correct recitation but cannot produce the correct answer, while around $12\%$ questions do not have the correction recitation but can be answered correctly anyway.

\section{Conclusion \& Discussion}
In this paper, we propose a novel recitation-augmented generation framework to improve language models' performance in the closed-book question-answering setting. We hypothesize that for knowledge-intensive NLP tasks, encouraging the model to explicitly recite a specific knowledge source would be helpful in augmenting its memory. In addition, we found that diversifying the recitation process can be beneficial as well since usually there exists multiple knowledge sources that could be used to answer the same question.
We show promising results over three large language models and across three different closed-book QA datasets, demonstrating the effectiveness of our proposed recite-and-answer approach.

One limitation of our method is that updating time-sensitive knowledge for a pure LLM-based method requires training or fine-tuning the LLMs on the new corpus, which can be costly. For future work, we plan to further validate the effectiveness of recitation-augmented generation for other knowledge-intensive NLP tasks in the closed-book setting, such as fact checking.

\section*{Acknowledgement}

We thank the support and feedback of many people from Google Brain team and the constructive suggestions from the anonymous reviewers.

\section*{Ethics Statement}

The goal of this paper is to use recitation as an intermediate step to generate more accurate factual knowledge in the model's outputs. Therefore, our method should in principle improve the faithfulness of the LLM systems. However, unlike retrieval-augmented generation (RAG) models that can use external trustworthy corpus, all the intermediate steps in RECITE is generated by the LLM itself, RECITE may further enhance the existing biases in the LLMs' model weights compared to RAG.

\section*{Reproducibility Statement}

\paragraph{Model weights} The model weights of two LLMs used in our experiments, i.e., UL2-20B \citep{tay2022unifying} and OPT-30B \citep{zhang2022opt}, are publicly released through GCP bucket (\url{gs://scenic-bucket/ul2}) and Github (\url{https://github.com/facebookresearch/metaseq}), respectively. The Codex (\texttt{code-davinci-002}) model is publicly available through API calls (\url{https://beta.openai.com/examples/}).

\paragraph{Evaluation datasets} The three evaluation datasets used in our experiments (Natural Questions\footnote{\url{https://github.com/google-research-datasets/natural-questions}}, TriviaQA\footnote{\url{https://nlp.cs.washington.edu/triviaqa/}}, and HotpotQA\footnote{\url{https://github.com/hotpotqa/hotpot}}) are all publicly accessible.

\paragraph{Prompts} We provide all the used prompts in the appendix.

\paragraph{Source code} Though the prompt examples in the appendix should be enough to reproduce all the results in our paper, we open-source all the evaluation code at \url{https://github.com/Edward-Sun/RECITE}.

\bibliography{iclr2023_conference}
\bibliographystyle{iclr2023_conference}

\newpage
\appendix
\section{Illustrations of Prompts and Language Model Outputs}

\begin{figure}[H]
    \centering
    \includegraphics[width=0.8\linewidth]{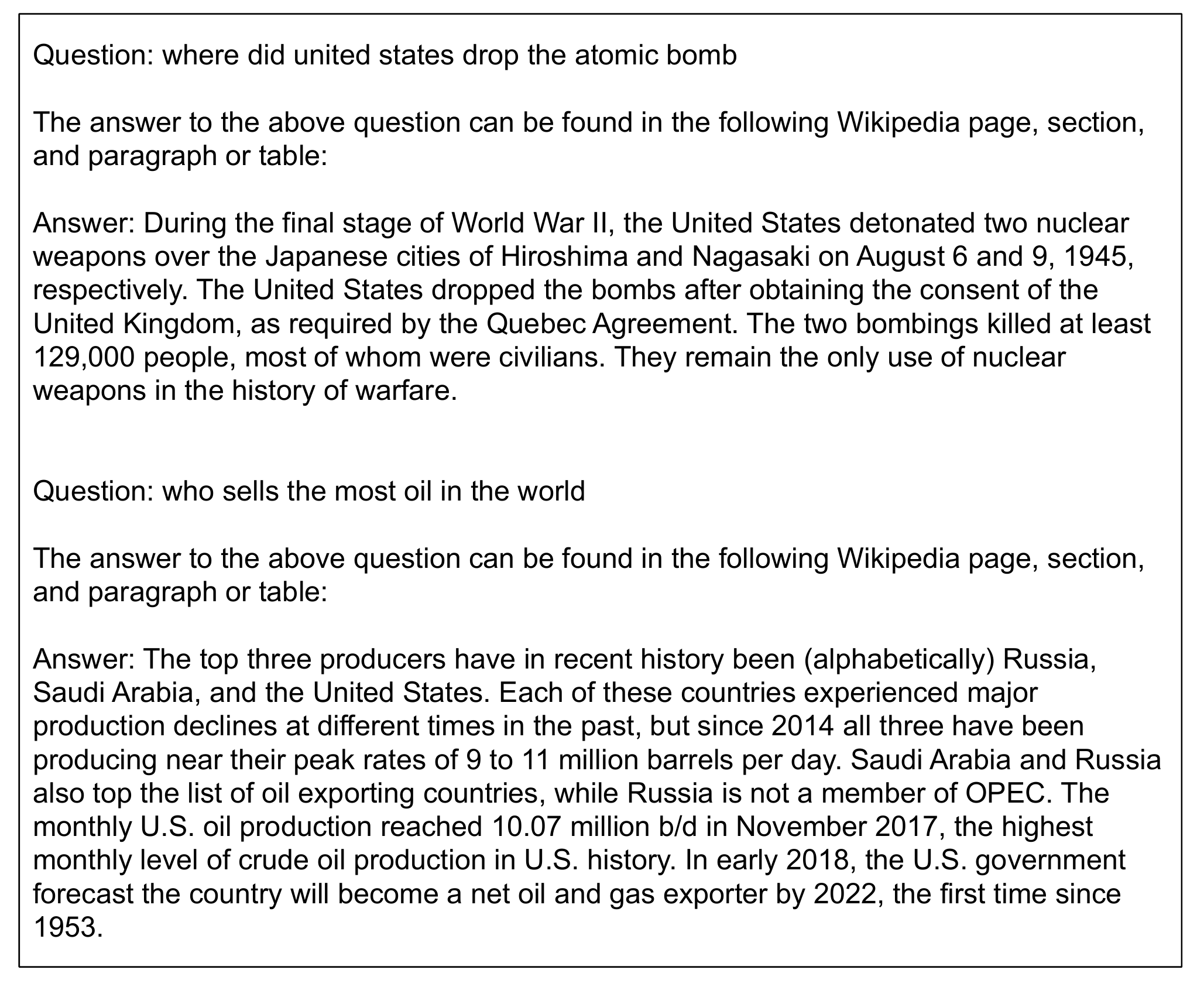}
    \caption{Two sampled evidence-recitation exemplars in Natural Questions (NQ) dataset. Notice that in NQ, we can directly use the ``long answer'' annotation as the recitation demonstrations.}
    \label{fig:app1}
\end{figure}

\begin{figure}[H]
    \centering
    \includegraphics[width=0.8\linewidth]{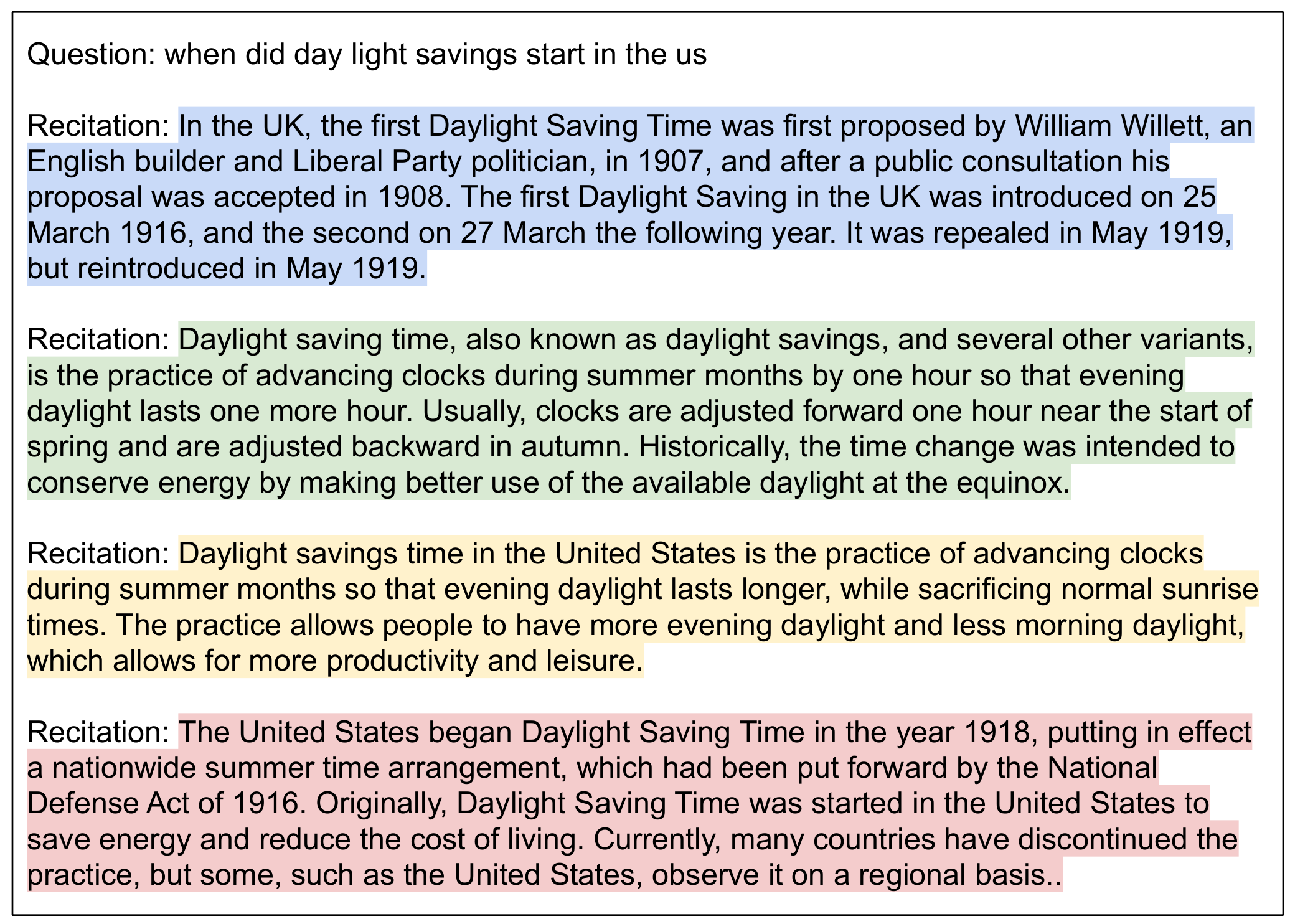}
    \caption{Four recitation passages sampled from UL2 for the same example question from Natural Questions dataset.}
    \label{fig:app2}
\end{figure}

\begin{figure}[H]
    \centering
    \includegraphics[width=0.8\linewidth]{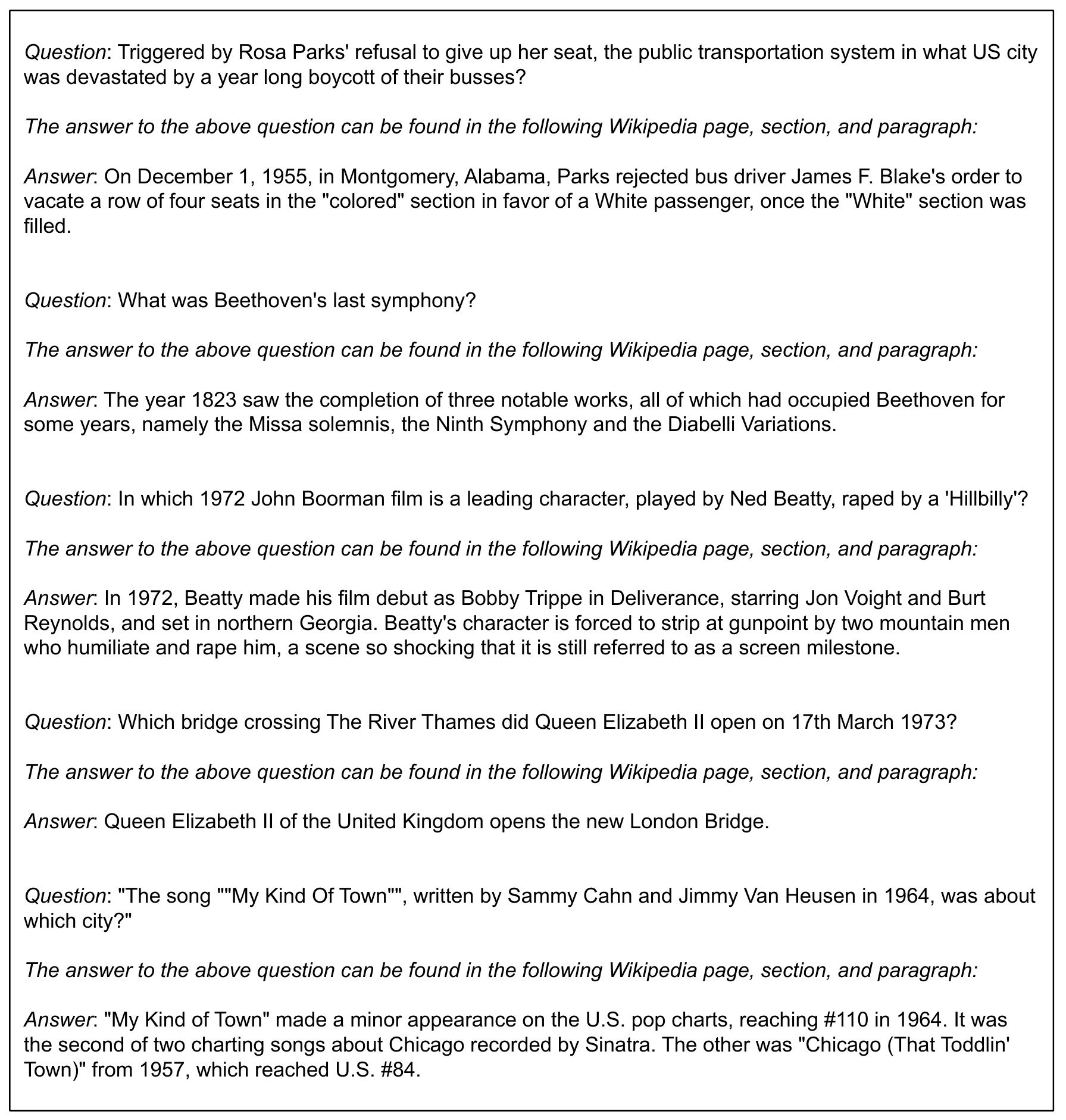}
    \caption{The 5-shot prompt we used for performing evidence-recitation on TriviaQA dataset.}
    \label{fig:app3}
\end{figure}

\begin{figure}[t]
    \centering
    \includegraphics[width=0.8\linewidth]{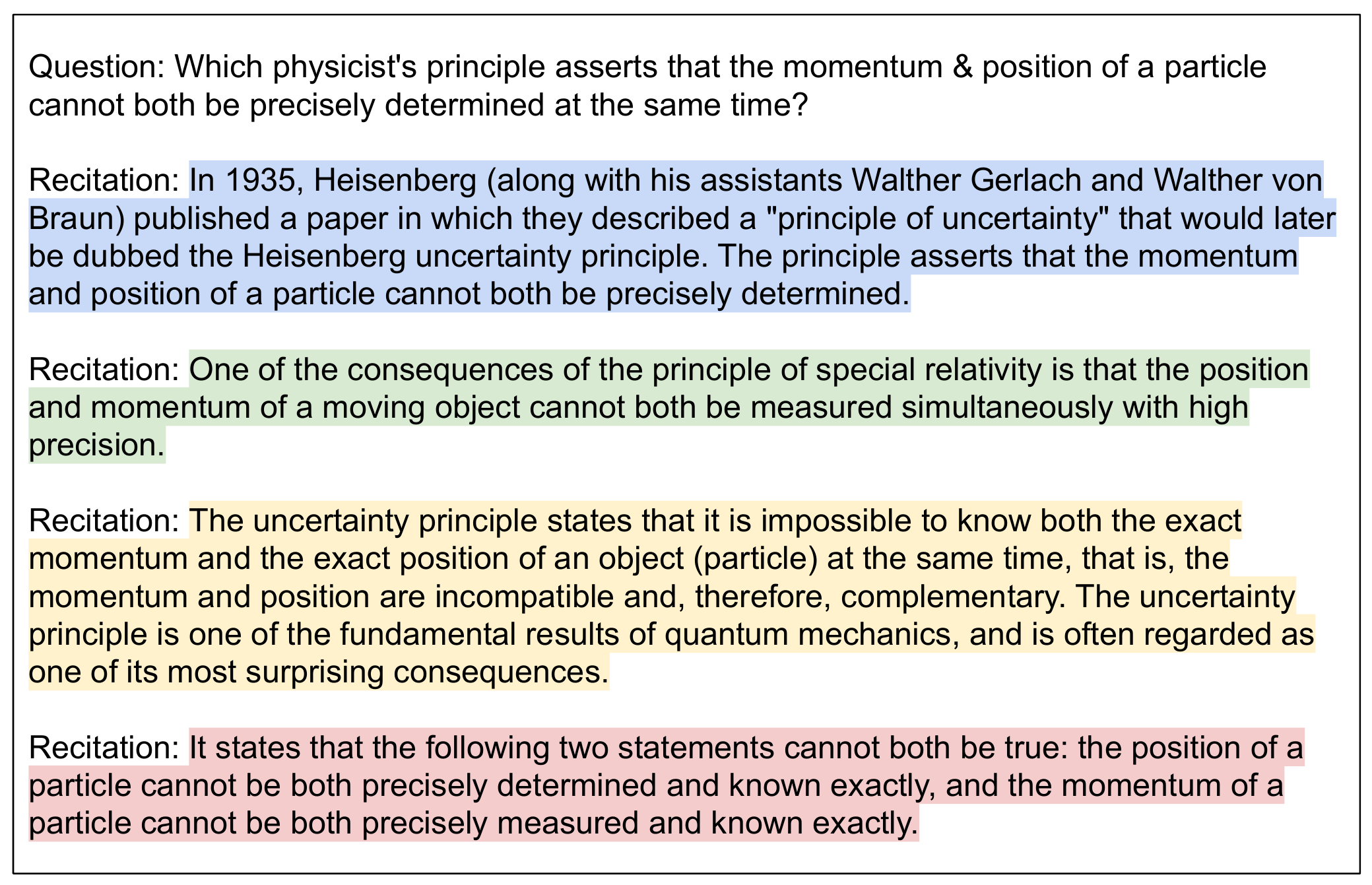}
    \caption{Four recitation passages sampled from UL2 for the same example question from TriviaQA dataset.}
    \label{fig:app4}
\end{figure}

\begin{figure}[t]
    \centering
    \includegraphics[width=0.7\linewidth]{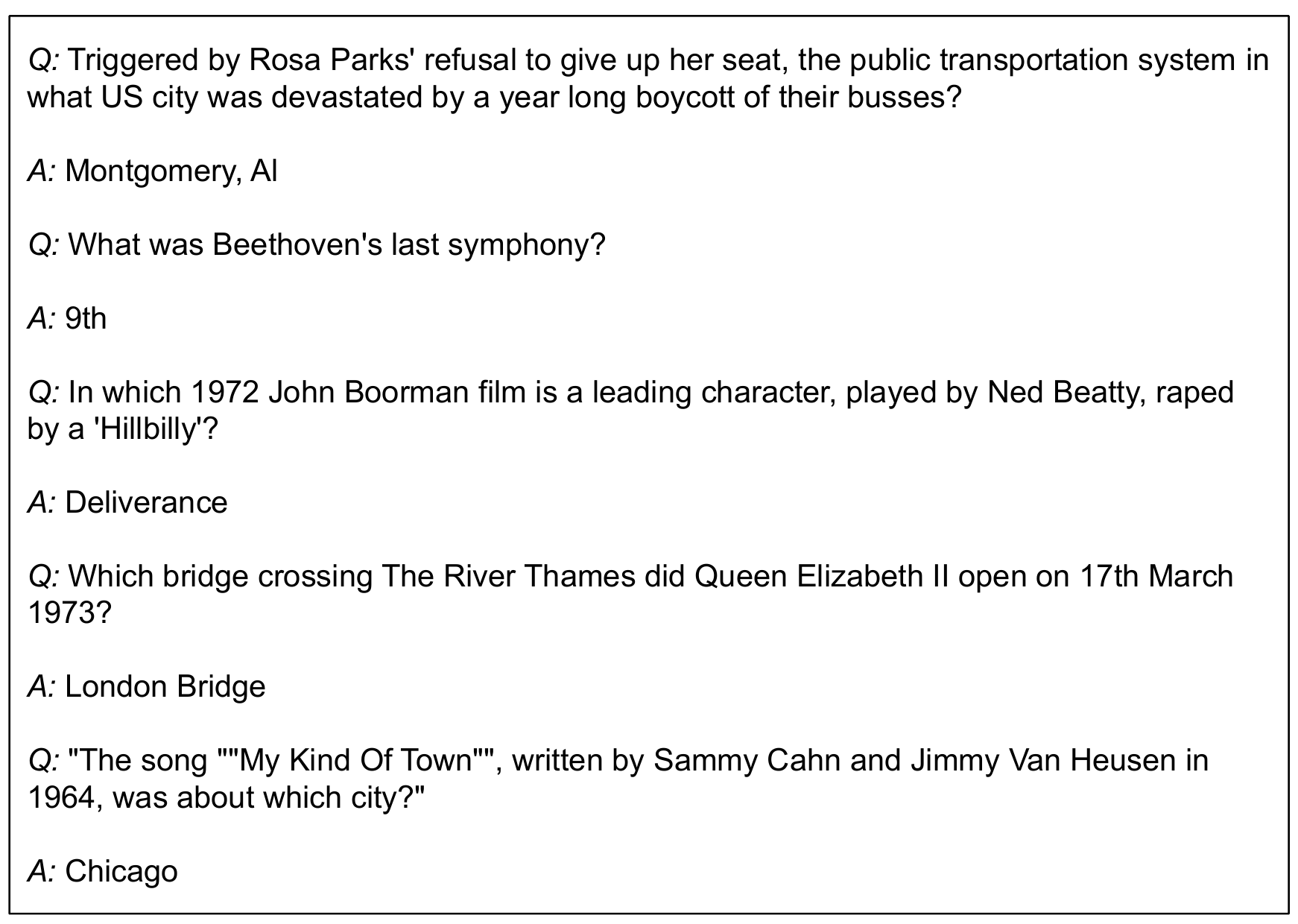}
    \caption{The 5-shot prompt we used for performing question-answering on TriviaQA dataset.}
    \label{fig:app5}
\end{figure}

\begin{figure}[t]
    \centering
    \includegraphics[width=1.0\linewidth]{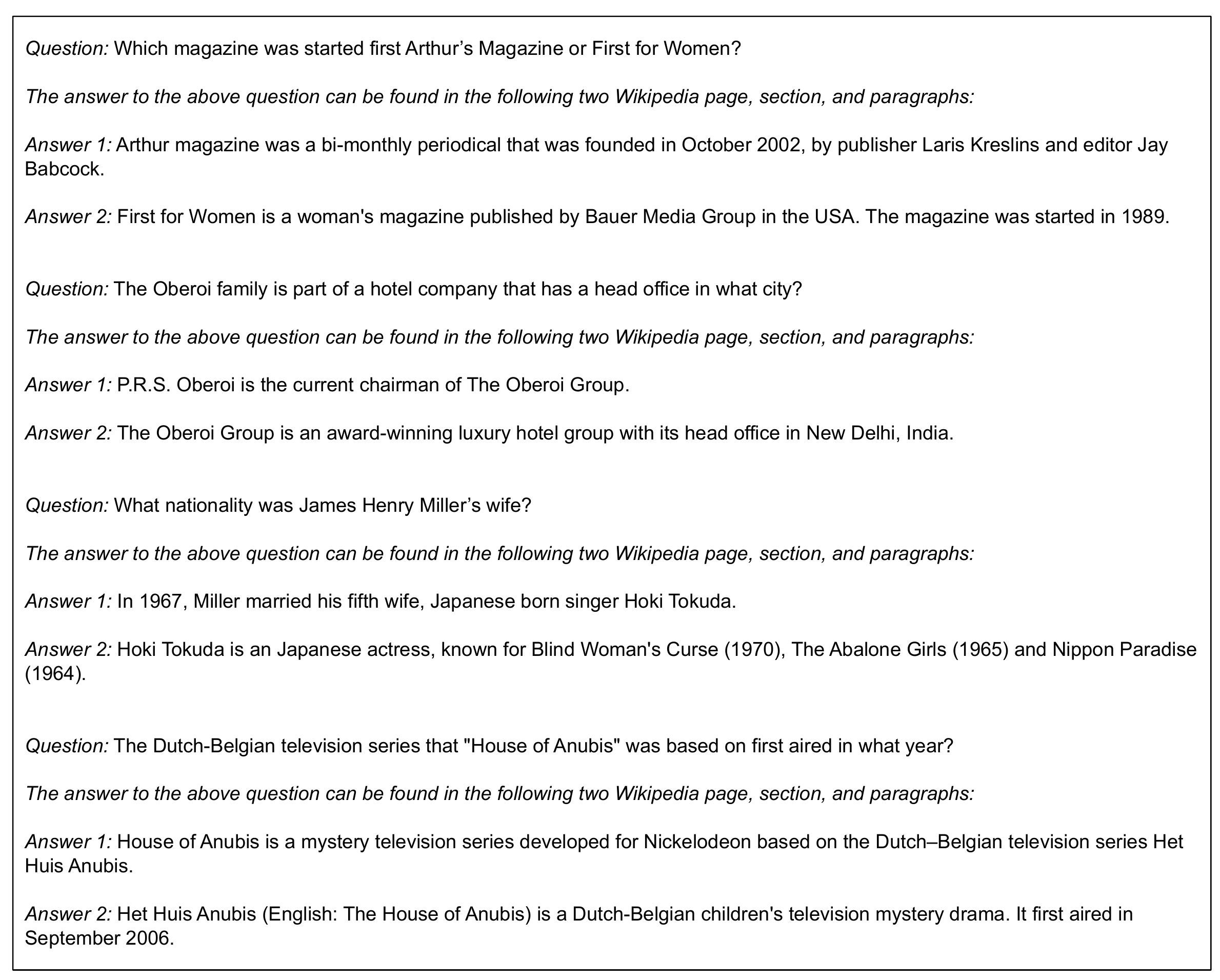}
    \caption{The 4-shot prompt we used for performing multiple-evidence recitation on HotpotQA dataset.}
    \label{fig:app6}
\end{figure}

\begin{figure}[t]
    \centering
    \includegraphics[width=0.8\linewidth]{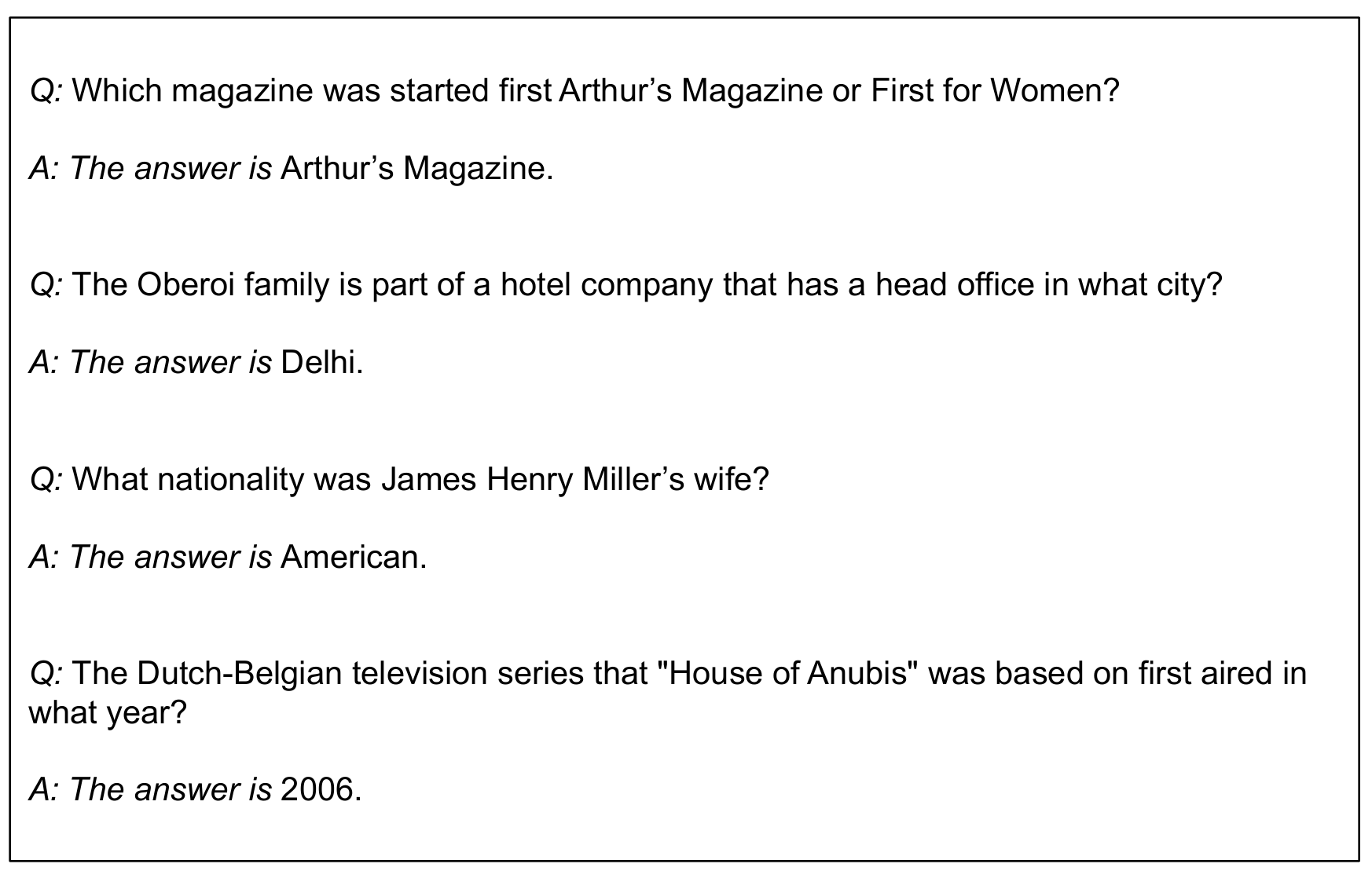}
    \caption{The 4-shot prompt we used for performing question-answering on HotpotQA dataset.}
    \label{fig:app7}
\end{figure}

\begin{figure}[t]
    \centering
    \includegraphics[width=0.8\linewidth]{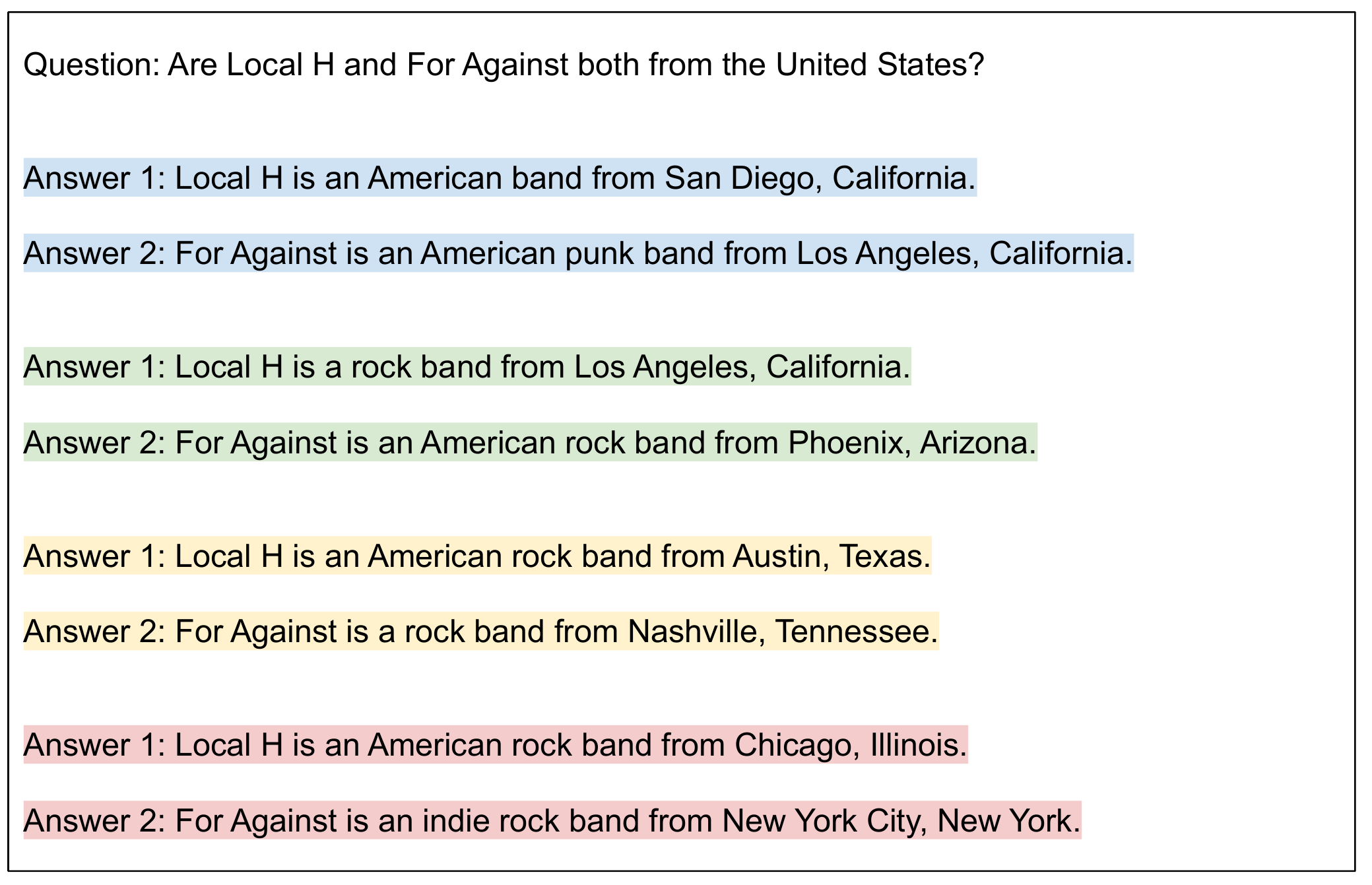}
    \caption{Four recitation sampled from UL2 for the same example question from HotpotQA dataset.}
    \label{fig:app8}
\end{figure}

\begin{figure}[t]
    \centering
    \includegraphics[width=0.8\linewidth]{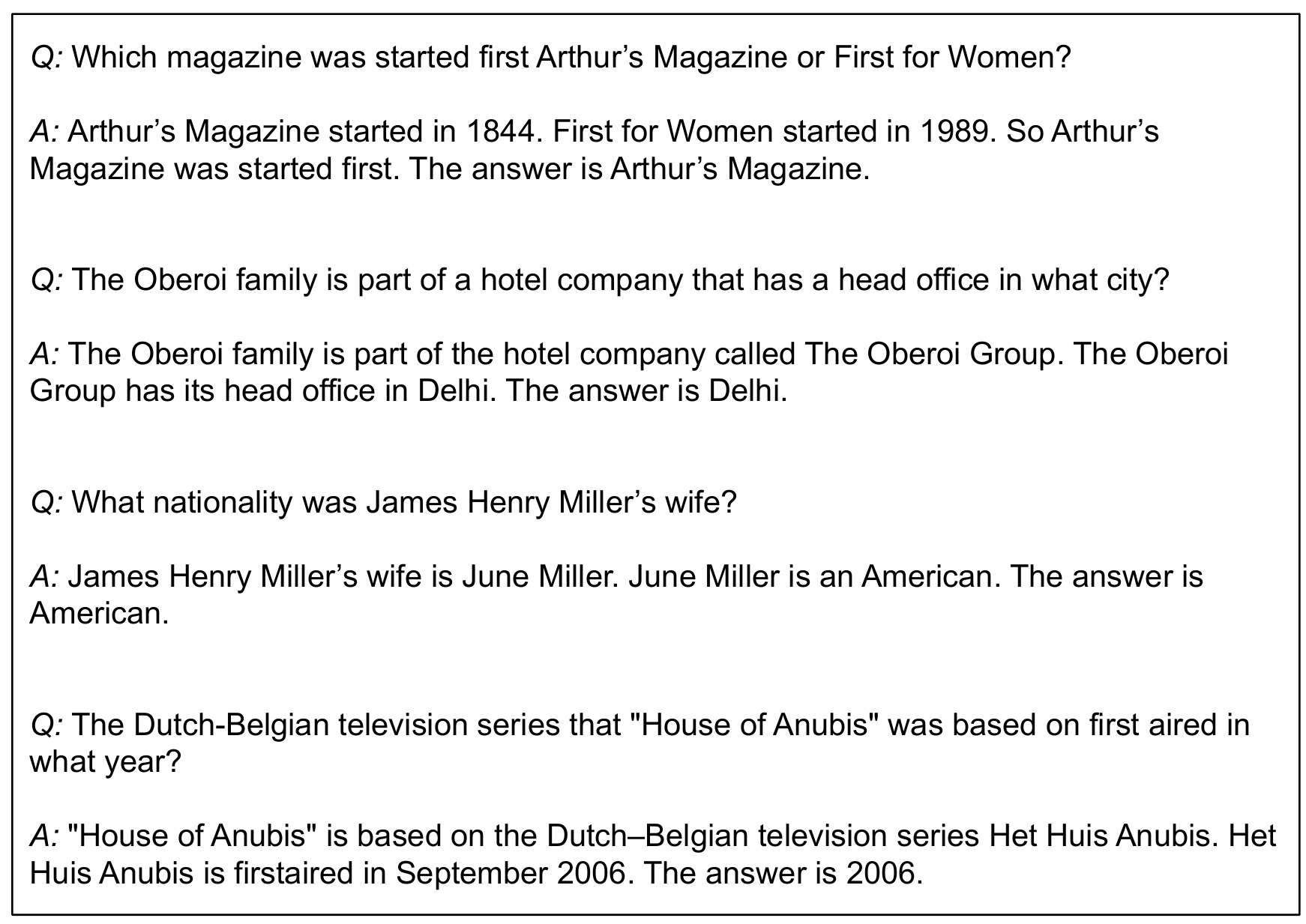}
    \caption{The 4-shot prompt we used for the Chain-of-Thought \citep{wei2022chain} baseline on HotpotQA dataset. The prompt is taken from \citep{wang2022rationale}.}
    \label{fig:app9}
\end{figure}

Fig.~\ref{fig:app1}-\ref{fig:app9} illustrate the evidence-recitation, question-answering prompt that we used for Natural Questions, TriviaQA, and HotpotQA dataset. We also provide the example sampled recitations for these datasets. Notice that for Natural Questions, we use the ``long answer'' annotation as the recitation in the prompt, while for the other two datasets, we manually compose a few recitation passages based on web search.

\section{Principles of prompt designs}

We mainly follow \citet{chowdhery2022palm} and use two new line symbols ``$\mathrm{\backslash n \backslash n}$'' as the separator between different components within exemplars, and use three new line symbol ``$\mathrm{\backslash n\backslash n\backslash n}$'' as the separator between different exemplars.

For the UL2 \citep{tay2022unifying} model, since its original SentencePiece \citep{kudo2018sentencepiece} vocabulary does not encode the new line symbol ``$\mathrm{\backslash n}$'', we instead use `` $\mathrm{;}$ '' to replace ``$\mathrm{\backslash n}$'' as the separator in all the prompts.

\section{Details of Passage Hint-based Fine-tuning}

For Natural Questions \citep{kwiatkowski2019natural} dataset, we assume that we have top-retrieved Wikipedia pages for the unannotated queries issued to the Google search engine by multiple user. We collect the passages in these pages as a corpus, and use the rule to annotate the hints of these passages.

To make a fair comparison with prompting-based models in both 5-shot and 64-shot, we only use 5 paired ``long answer''-question exemplars as the prompt to generate the synthetic question for the sampled passages from the Wikipedia hint-passage corpus, and thus construct the synthetic question-hint-passage paired fine-tuning data.

We train \palm~in the constructed corpus for 10,000 steps with a batch size of 64, which takes approximately 1 day in 64 TPUv4 chips\footnote{\url{https://cloud.google.com/tpu/docs/v4-users-guide}}. The fine-tuned model can be used for passage hint-diversified recitation without any further prompts.

\section{Pre-trained Language Models} \label{sec:lms}

\paragraph{\palm}
\palm~
is a family of densely activated decoder trained on the language modeling objective. It has strong capabilities in in-context few-shot learning, multilingual, as well as reasoning tasks. In this paper, we use the \palm~model with 62B parameters.

\paragraph{UL2} UL2 (Unifying Language Learning, \citealt{tay2022unifying}) is an encoder-decoder model trained on a mixture of denoising tasks in a unified framework. In this paper, since we mainly focus on the in-context learning ability of language models, we use UL2-20B in the S-Denoiser mode (i.e., pre-trained with the prefix language modeling)\footnote{This can be achieved by append the ``[NLG]'' and  ``[extra\_id\_0]'' token to the beginning and the end of the prompt.}.

\paragraph{OPT} OPT (Open Pre-trained Transformer language model, \citealt{zhang2022opt}) is a family of recently released open-source densely activated language model that aims to re-reproduce comparable results as GPT-3 \citep{brown2020language}. We use the 30B one\footnote{\url{https://github.com/facebookresearch/metaseq}} in this paper.

\paragraph{Codex} Codex \citep{ouyang2022training,chen2021evaluating} is a variant of GPT-3 model \citep{brown2020language} that can understand code. We use the public OpenAI API\footnote{\url{https://openai.com/api/}} to access the conditional generation outputs of the \texttt{code-davinvi-002} model in this paper.

\begin{figure}[H]
\centering
\begin{minipage}{.5\textwidth}
  \centering
  \includegraphics[width=.9\linewidth]{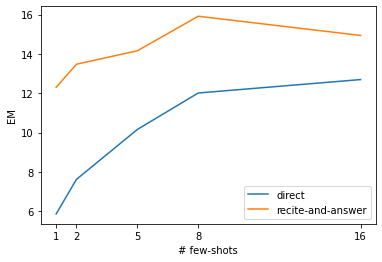}
\end{minipage}%
\hfill
\begin{minipage}{.5\textwidth}
  \centering
  \includegraphics[width=.9\linewidth]{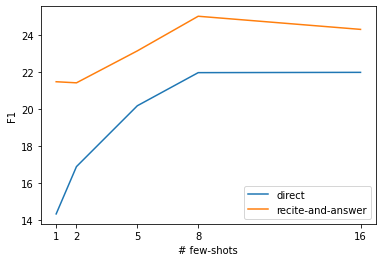}
\end{minipage}
\vspace{-0.1in}
\caption{Nature Questions EM/F1 on UL2-20B with different \# of shots.}
\label{tab:analysis_nq_n_shots}
\end{figure}

\begin{figure}[H]
\centering
\begin{minipage}{.5\textwidth}
  \centering
  \includegraphics[width=.9\linewidth]{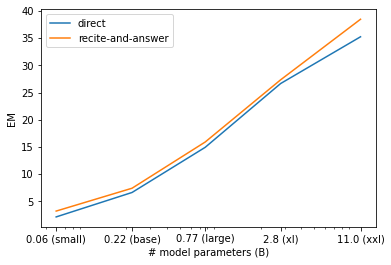}
\end{minipage}%
\hfill
\begin{minipage}{.5\textwidth}
  \centering
  \includegraphics[width=.9\linewidth]{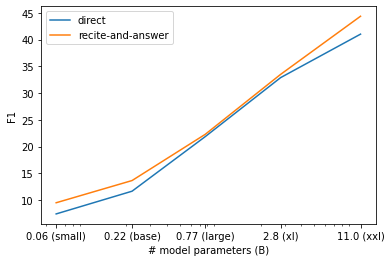}
\end{minipage}
\vspace{-0.1in}
\caption{TriviaQA EM/F1 on 5-shot FLAN-T5 with different model sizes.}
\label{tab:analysis_trivia_size}
\end{figure}

\begin{center}
\begin{minipage}{1.0\linewidth}
\begin{algorithm}[H]
\caption{Per-question Error Analysis}
\begin{algorithmic}
\Require{ $N$ ground-truth answer labels $\{A_i\}_{i=1}^{N}$ for a single question}
\Require{ recitations $\{R_i\}_{i=1}^{K}$ and answer predictions $\{P_i\}_{i=1}^{K}$ from $K$ paths}
\State $\mathrm{normalized\_answers}$ $\gets$ $\{\mathrm{normalize}(A_i)\}_{i=1}^N$
\If{ $\mathrm{normalize}
(\mathrm{majority\_vote}(\{P_i\}_{i=1}^{K}))
\in \mathrm{normalized\_answers}$}
    \State \Return{``$\mathrm{Hits@Majority}$''}
\ElsIf{$\mathrm{Any}(\{\mathrm{normalize}
(P_i)
\in \mathrm{normalized\_answers}\}_{i=1}^{K})$}
    \State \Return{``$\mathrm{Hits@20-Path.}$''}
\ElsIf{$\mathrm{Any}(\{\mathrm{Any}(\{\mathrm{normalized\_answers}_j \in \mathrm{normalize}(R_i)\}_{i=1}^{K})\}_{j=1}^N)$}
    \State \Return{``$\mathrm{Hits@20-Recit.}$''}
\Else
    \State \Return{``$\mathrm{No Recit.}$''}
\EndIf
\end{algorithmic}
\label{alg:error}
\end{algorithm}
\end{minipage}
\end{center}

\section{Analysis of the number of examples in few-shot learning}

We analyze the influence of the number of shots on Natural Questions dataset for standard-prompting model and our recite-and-answer model in Fig.~\ref{tab:analysis_nq_n_shots}. We can see that recite-and-answer prompting achieves consistent improvement over standard prompting, while the largest improvement is achieved in the 1-shot setting.

\section{Analysis of model sizes on instruction-finetuned language models}

Instruction-finetuned language models \citep{sanh2021multitask,bach2022promptsource,wei2021finetuned,chung2022scaling} trained on a collection of datasets phrased as instructions has been shown to improve model performance and generalization to unseen tasks. They tend to show better zero-shot or few-shot performance under the same model sizes compared to their vanilla LM pre-trained counterparts. We analyze the performance on one representative instruction-finetuned language model, FLAN-T5 \citep{chung2022scaling}, on TriviaQA dataset for standard-prompting model and our recite-and-answer model in Fig.~\ref{tab:analysis_trivia_size}. We can see that recite-and-answer prompting achieves consistent improvement over standard prompting, while the largest improvement is achieved in the largest ``xxl (11B)'' setting.

\section{Language Model Hyperparameters}

We report the hyperparameters of the LLMs we used in Tab.~\ref{tab:lm_hyper}.

\begin{table}[t]
\caption{Robustness evaluation of UL2 on \mbox{TriviaQA} with different few-shot exemplars over 5 random seeds.}
    \centering
    \small
    \begin{tabular}{l c c}
    \toprule
    & EM\textsubscript{(5)}\ & F1\textsubscript{(5)}\\
    \midrule
    Standard (direct)& 48.42 ($\pm$ 0.71) & 53.85 ($\pm$ 0.57)\\
    RECITE (5-path) & 49.75 ($\pm$ 0.50) & 54.78 ($\pm$ 0.46)\\
    RECITE (20-path) & \textbf{52.68} ($\pm$ 0.62) & \textbf{58.05} ($\pm$ 0.58)\\
    \bottomrule
    \end{tabular}
    \vspace{-0.05in}
    \label{tab:robustness_evaluation}
\end{table}

\begin{table}[H]
    \centering
    \small
    \caption{The model hyper-parameters of the large language models used in our experiments. Note that for Codex (i.e., \texttt{code-davinci-002}), the details (including size) are unknown, so we report the hyperparameters of GPT-3 as a common educational guess.}
    \begin{tabular}{l c c c c c c}
    \toprule
         Model & Type & \# of Layers & \# of Heads & $d_{\mathrm{model}}$ & \# of Parameters (B) \\
    \midrule
         \palm-62B & decoder-only & 64 & 32 & 8192 & 62.50 \\
         UL2-20B & encoder-dencoder & 32 / 32 & 16 & 4096 & 19.46 \\
         OPT-30B & decoder-only & 48 & 56 & 7168 & 29.97 \\
         Codex$^*$ & decoder-only & 96 & 96 & 12288 & 175.0 \\
    \bottomrule
    \end{tabular}
    \label{tab:lm_hyper}
\end{table}

\begin{figure}[t]
    \centering
    \includegraphics[width=\linewidth]{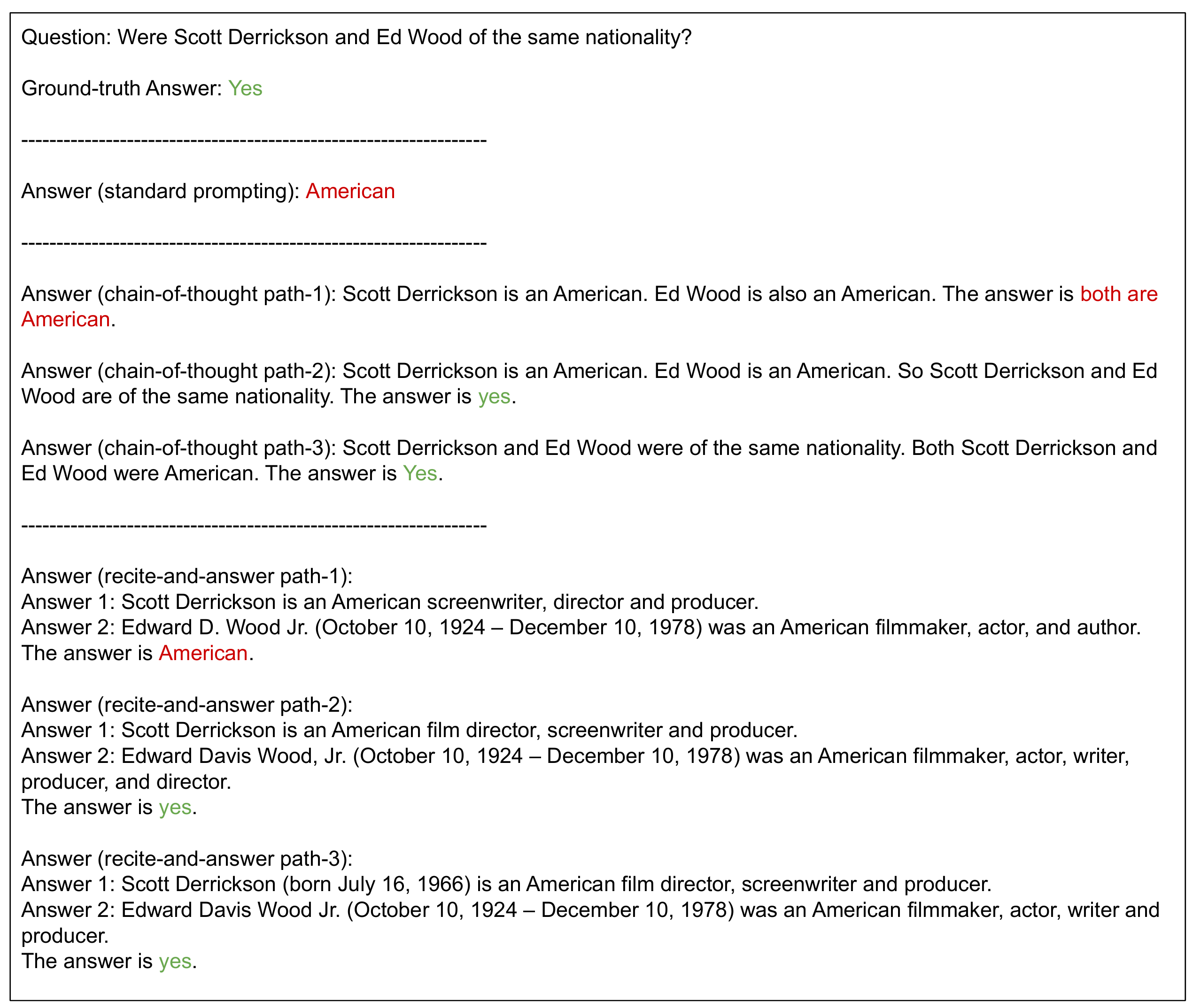}
    \caption{Qualitative comparisons between standard prompting, chain-of-thought, and recite-and-answer on HotpotQA evaluation example (I).}
    \label{fig:qual_hq_1}
\end{figure}

\begin{figure}[t]
    \centering
    \includegraphics[width=\linewidth]{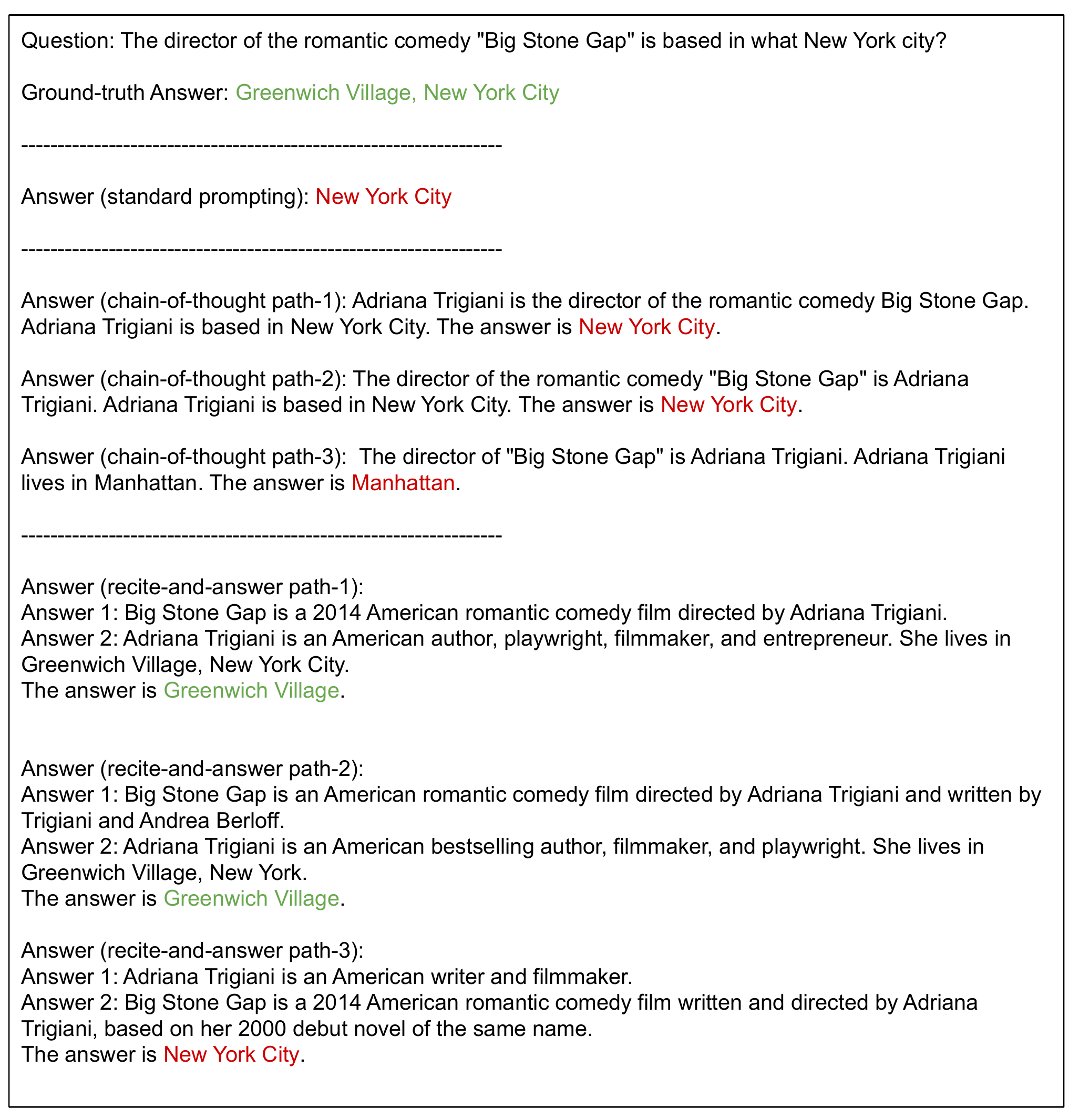}
    \caption{Qualitative comparisons between standard prompting, chain-of-thought, and recite-and-answer on HotpotQA evaluation example (II).}
    \label{fig:qual_hq_2}
\end{figure}

\end{document}